\documentclass[twocolumn]{article}
\usepackage[a4paper,margin=0.7in]{geometry}
\usepackage{graphicx}
\usepackage{booktabs}
\usepackage{tabularx}
\usepackage{adjustbox}
\usepackage{pdflscape}
\usepackage{setspace}
\usepackage{rotating}
\usepackage{makecell} 
\usepackage{orcidlink}
\usepackage{cite}
\usepackage{amsmath,amssymb,amsfonts}
\usepackage{algorithmic}
\usepackage{algorithm,algorithmic}
\usepackage{hyperref}
\usepackage{textcomp}
\usepackage{listings}
\usepackage{fvextra}
\usepackage{hyperref}
\usepackage{rotating,fancyhdr,ifthen}
\usepackage{float}
\usepackage{afterpage}
\usepackage{threeparttable}
\usepackage{float}
\usepackage{afterpage}

\usepackage{threeparttable}

\title{Can Large Language Models Reliably Extract Physiology Index Values from Coronary Angiography Reports?}

\author{Sofia Morgado\orcidlink{0009-0000-7028-4430}, Filipa Valdeira\orcidlink{0000-0003-1290-4798}, Niklas Sander\orcidlink{0009-0000-7310-9056}, Diogo Ferreira \orcidlink{0009-0005-0841-5600}, Marta Vilela\orcidlink{0009-0009-4359-6565},\\ Miguel Menezes\orcidlink{0000-0001-8363-0363}, and Cl\'audia Soares\orcidlink{0000-0003-3071-6627}
\thanks{Received on \today. 
This work is funded by national funds through FCT – Fundação para a Ciência e a Tecnologia, I.P., in the scope of project HyCARE (2024.07361.IACDC), Laboratory for Computer Science and Informatics, NOVA LINCS  (UID/04516/2025), the Phd Scholarship (2025.04592.BDANA) and Center for Computational and Stochastic Mathematics (UIDB/04621/2025).} 
\thanks{Sofia Morgado is with the Computer Science Department at Nova School of Science and Technology, Universidade NOVA de Lisboa, Caparica, Portugal (FCT-UNL) - +351960047437 (\href{mailto:sv.morgado@campus.fct.unl.pt}{sv.morgado@campus.fct.unl.pt} ).}
\thanks{Filipa Valdeira is with Center for Computational and Stochastic Mathematics - Department of Mathematics at Instituto Superior Técnico (\href{mailto:filipa.valdeira@tecnico.ulisboa.pt}{filipa.valdeira@tecnico.ulisboa.pt}).}
\thanks{Niklas Sander is with the Department of Computer Science at FCT-UNL (\href{mailto:n.sander@campus.fct.unl.pt}{n.sander@campus.fct.unl.pt}).}
\thanks{Diogo Ferreira is with the Cardiology Department at Centro Hospitalar Universit\'ario Lisboa Norte (CHULN) (\href{mailto:28833@chln.min-saude.pt}{28833@chln.min-saude.pt}).}
\thanks{Marta Vilela is with the Cardiology Department at CHULN (\href{mailto:28859@chln.min-saude.pt}{28859@chln.min-saude.pt}).}
\thanks{Miguel Menezes is with the Cardiology Department of CHULN and Faculdade de Medicina da Universidade de Lisboa (\href{mailto:miguel.menezes@ulssm.min-saude.pt}{miguel.menezes@ulssm.min-saude.pt} ).}
\thanks{Cl\'audia Soares is with Nova LINCS - Department of Computer Science at FCT-UNL (\href{mailto:claudia.soares@fct.unl.pt}{claudia.soares@fct.unl.pt}).}
}

\begin{document}

\maketitle

\begin{abstract}
Coronary angiography (CAG) reports contain clinically relevant physiological measurements, yet this information is typically in the form of unstructured natural language, limiting its use in research. We investigate the use of Large Language Models (LLMs) to automatically extract these values, along with their anatomical locations, from Portuguese CAG reports. To our knowledge, this  study is the first addressing physiology indexes extraction from a large (1342 reports) corpus of CAG reports, and one of the few focusing on CAG or  Portuguese clinical text.

We explore local privacy-preserving general-purpose and medical LLMs under different settings. Prompting strategies included zero-shot, few-shot, and few-shot prompting with implausible examples. In addition, we apply constrained generation and introduce a post-processing step based on RegEx. 
Given the sparsity of measurements, we propose a multi-stage evaluation framework separating format validity, value detection, and value correctness, while accounting for asymmetric clinical error costs.

This study demonstrates the potential of LLMs in for extracting physiological indices from Portuguese CAG reports. Non-medical models performed similarly, the best results were obtained with Llama with a zero-shot prompting, while GPT-OSS demonstrated the highest robustness to changes in the prompts. While MedGemma demonstrated similar results to non-medical models, MedLlama's results were out-of-format in the unconstrained setting, and had a significant lower performance in the constrained one. Changes in the prompt techinique and adding a RegEx layer showed no significant improvement across models, while using constrained generation decreased performance, although having the benefit of allowing the usage of specific models that are not able to conform with the templates.
\textbf{Project Repository}: \url{https://github.com/sofiabmorgado/cag-llm-extraction}
\end{abstract}

\section{Introduction}
Coronary angiography (CAG) is the cornerstone in the management of coronary artery disease (CAD), the leading cause of mortality globally \cite{Who2025}. It offers detailed visualizations of coronary anatomy and enables measurements of physiology indices, the fractional flow reserve (FFR) and instantaneous wave-free ratio (iFR). These are pressure-based lesion-specific indices, measured only in vessels where intermediate lesions were identified, and indicate the impact of coronary stenosis on myocardial perfusion \cite{Pijls1995}, thus are crucial step for the decision to procede with angioplasty, a treatment intervention that can significantly improve the prognosis in these patients. A detailed report of CAG documents the findings, measurements, and outcomes of the angiography and angioplasty \cite{Vrints2024}.

Large scale scientific research is pivotal in the development of comprehensive diagnostic approaches and precise treatment strategies to improve prognosis of CAD. Scientific research has comprised a range of tasks in the field, such as image segmentation \cite{Menezes2023} or event prediction \cite{Wang2024b}.
Yet, the interpretation of CAG findings often reside in unstructured natural language reports, 
resulting in an expensive and labor-intensive manual extraction of significant information, constraining prospects for extensive analysis and research based on this data.

Large language models (LLMs) have demonstrated remarkable state-of-the-art (SOTA) capabilities in understanding and generating natural language and have been used in the medical context for clinical, research and educational applications \cite{Thirunavukarasu2023}
Particularly, SOTA models have reached levels that are comparable to those of human experts in the medical field \cite{Veen2024,Singhal2023, Kanzawa2024}. Its usage spans a wide array of tasks, such as question-answering, clinical reasoning, eletronic health records summarization or multimodal medical image analysis \cite{Thirunavukarasu2023, Wang2024}; and multiple specialties, from psychiatry \cite{Omar2024} to radiology \cite{Woznicki2024}, and cardiology \cite{Boonstra2024}.
Leveraging LLMs to transform unstructured medical data into structured actionable data is a promising avenue \cite{Kanzawa2024, Boonstra2024}.
However, there are strict requirements to the use of LLMs in healthcare, such as the need for locally-hosted privacy-preserving solutions, credible methods with very limited risk of hallucinations \cite{Boonstra2024,Mccoy2024,Thirunavukarasu2023} and explainability, and trustworthiness \cite{Wang2024, Quer2024}.

There are limited studies on the use of LLMs specifically in cardiology, and the few that exist are constrained by small sample sizes and require further external validation. To our knowledge, only one study has investigated the use of LLMs for the extraction of information (IE) from angiography reports; however, it relied on a very small corpus and used proprietary models \cite{Song2026}, thus not guaranteeing crucial privacy preservation measures in healthcare. Moreover, we found no studies addressing the extraction of physiological indices from CAG, and none conducted in Portuguese, a setting that may pose challenges for predominantly English-trained models.

\subsection{Objectives}
This project aims to investigate the potential of LLMs to extract physiology indexes from Portuguese CAG reports, addressing current limitations related to unstructured clinical documentation. 
This project will empirically evaluate privacy-preserving LLM-based methods for the extraction of the numerical values of FFR and iFR, and their anatomical localization, and further aims at

\begin{itemize}
\item Evaluate the feasibility and performance of LLMs in the extraction of FFR and iFR values and their anatomical localization from natural language CAG reports;
\item Assess the performance of LLMs for IE under different prompting, constraint and post-processing strategies;
\item And compare the performance of general purpose and medical LLMs for clinical IE.
\end{itemize}

\subsection{Contributions}
Our main contributions include:

\begin{itemize}
\item \textbf{The application to a novel clinical and language setting}. To our knowledge, we present the first study applying LLMs for physiology indices extraction from CAG reports, evaluating the performance in Portuguese clinical text, while guaranteeing privacy-preserving methods. 
\item \textbf{A comprehensive empirical ablation study} for improving structured extraction, including few-shot prompting using implausible values, prompt robustness evaluation, constrained generation and a post-processing RegEx based layer. 
\item \textbf{A comparison of medical and general-domain LLMs}, showing that while medical LLMs might have greater clinical reasoning, they were not as useful for IE as general purpose models.
\item \textbf{An IE evaluation framework for sparse clinical extraction tasks}, a multi-stage evaluation pipeline tailored to settings with high proportions of missing values and asymmetric error costs, where hallucinations of values that are not present are more severe errors than missing the extraction of a given value.
\end{itemize}

\section{Related Work}
The latest general purpose LLMs, including proprietary models, such as GPT-5 \cite{gpt5}, 
and open-source models such as Llama3 \cite{llama3}, Mistral \cite{Jiang2023} and DeepSeek-R1 \cite{deepseek}, 
have demonstrated proficiency in a wide range of tasks. Some of these models have been applied to medical tasks \cite{Eriksen2023}, such as medical reasoning (GPT4) \cite{Omar2024}, treatment guidance (GPT4) \cite{Liu2023}, and structuring medical data (Llama2) \cite{Woznicki2024, Adams2023}. 
Furthermore, some limitations have been found when applying general purpose models for medical specific tasks, such as missing important diagnosis \cite{Eriksen2023} and the presence of hallucinations \cite{Liu2023}. For that reason, there has been a large effort to produce specialized models.
Medical LLMs include  MedGemma \cite{Medgemma}, MedLlama \cite{Medllama}, MedPalm \cite{Singhal2023}, MedBERT \cite{Rasmy2021}, GatorTron \cite{Yang2022}, and MedGemini \cite{Saab2024}, as well as
 MediAlbertina \cite{Nunes2024} and BioBERTpt \cite{Schneider2020}, available in portuguese.

 This work focuses on numerical extraction with generative large language models. Encoder-only models, including BERT-based models, require additional engineering to be suitable for numerical extraction, and the studied prompting- and decoding-approaches for generative LLMs are not applicable. Models that do not support prompting-based extraction and generative generation of JSON are therefore not considered in this work. 

 
Specialized medical models are finetuned for a specific medical task, such as Llama3 with MedPALM dataset for an assistant medical chatbot \cite{Patel2024} or T5 with a proprietary dataset to structure radiology reports \cite{Bergomi2024}.
While some studies indicate that fine-tuning might improve the model's performance and decrease hallucinations \cite{Bergomi2024, Kanzawa2024}, others indicate effective prompt engineering, such as few-shot and chain-of-thought prompting, has demonstrated the capability to enhance generalist models for medical benchmarks, matching or exceeding the performance of fine-tuned models \cite{Maharjan2024}. 
Novel prompting strategies, such as the usage of Implausible values in the few-shot examples have shown improvements in the results \cite{Adam2025}.


\subsection{LLMs and Cardiology} 
The promissing results of LLMs have increased their usage in diverse tasks in cardiology in the last few years, including generation of medical notes, information extraction, risk prediction, and interpretation of diagnostic tests \cite{Boonstra2024, Quer2024, Bhattaru2024}. Nonetheless, their effectiveness continues to be inconsistent across diverse clinical settings \cite{Gendler2024}, underscoring the need for aditional evaluation and improvement \cite{Altamimi2024}.

Recent studies have compared the performance  of various LLMs in answering cardiology-related clinical questions, using OpenAI GPT \cite{Lee2023, Novak2024} and Google Bard \cite{Novak2024}, showing understanding of medical terminology, clinical context and knowlegde on the established guidelines.
Given their popularity and SOTA performance, GPT models are the most frequently used in the literature \cite{Harskamp2024,Sarraju2023,Dimitriadis2024, Riddell2023}. However, they have also provided incomplete, inconclusive, or inappropriate responses \cite{Harskamp2024}, although experts have considered their responses trustworthy and valuable for patient education \cite{Dimitriadis2024, Riddell2023}. Nevertheless, GPT models are mostly proprietary, raising concerns about data privacy, as they do not offer the same level of security as locally deployed models. However, lately OpenAI has released GPT-OSS \cite{GPT-OSS}, an open-source model with similar capabilities as their mini models.

In this context, a recent publication introduced CardioBERTpt, a set of pre-trained BERT-based models designed for cardiology tasks in Portuguese from Brazil fine-tuned for named-entity recognition. These models achieved SOTA results for the analyzed corpora, demonstrating that a large volume of representative clinical data can significantly enhance performance in medical NLP tasks \cite{Schneider2023}. Nevertheless, they were optimized for a task different from ours, which restricts their applicability to the objectives of this project.

Notwithstanding the promise of LLMs in cardiology, the European Society of Cardiology has expressed concerns about the utilization of LLMs in cardiology, emphasizing the necessity for models that physicians can trust, that provide patient benefits, and that ensure safety \cite{Boonstra2024}.
There is the need for privacy-preserving techniques \cite{Mccoy2024} and rigorous bias evaluation and mitigation \cite{Quer2024}. In most studies, ethical concerns such as data privacy were not adequately addressed. A key limitation of publications using GPT models is that the models used are proprietary, and any data processed is shared with OpenAI, raising potential confidentiality issues. 
Additionally, while these studies demonstrate the extent of LLMs' knowledge in cardiology, the tasks evaluated differ significantly from the specific application we aim to explore. 
Notably, we found no studies focusing on the use of LLMs for structuring coronary angiography reports in portuguese.

\subsection{Medical Report Structuring}
The task of automatically structuring natural language from medical exam reports has been explored in several studies, compared in Table \ref{tab:comparison}. 

\begin{table*}[htbp]
    \centering
    \footnotesize
    \caption{Comparison of studies on report structuring for medical diagnostic methods with LLMs}
    \begin{tabular}{l*6c}\toprule
        \textbf{Ref} & \textbf{Models} & \textbf{Privacy} & \textbf{Language} &  \textbf{Domain} & \textbf{Extracted Data} &  \textbf{Sample size} \\ \midrule
        \cite{Woznicki2024} & Llama2 & Yes & En, De  & Chest X-ray  & \makecell{Closed-ended questions\\ (Multichoice)}   & 399   \\ 
        \hline
        \cite{Adams2023} & GPT-4 & No & En, De  &  \makecell{Chest X-ray, \\ CT-scan, MRI } & Free-text  & 923   \\ 
        \hline
        \cite{Mukherjee2023} & Vicuna & Yes & En  & Chest X-ray   & \makecell{Closed-ended questions\\ (Yes/No)}  & 100 \\ 
        \hline
        \cite{Bergomi2024} & T5 (fine-tuned) & Yes & En & CT scan & \makecell{Free-text, Multichoice \\ Numerical} & 174  \\ 
        \hline
        \cite{Song2026} & \makecell{GPT-4o, Gemini, \\ Claude}  & No & En & CAG & \makecell{Closed-ended questions\\ (Multichoice or Numerical)} & 250   \\ 
        \hline
        \textbf{Ours} & \makecell{Mistral, LLama3, \\ GPT-OSS, \\ MedLlama2, MedGemma}  & Yes & Pt & CAG & Numerical (FFR/iFR)  & 1342  \\ 
        \hline
    \end{tabular}
    \begin{tablenotes}
    \centering
    \item De - German; En - English.
    \end{tablenotes}
    \label{tab:comparison}
\end{table*}

Most studies have primarily focused on chest X-ray, CT and MRI scan reports, written both in English and German, with language models such as Llama2 \cite{Woznicki2024}, GPT-3 \cite{Adams2023}, and Vicuna \cite{Mukherjee2023} being utilized without further fine-tuning. The combination of template structures, such as JSON templates  \cite{Woznicki2024, Adams2023, Mukherjee2023}, and libraries for output control (Guidance) \cite{Woznicki2024}, have been used to retrieve simple information, such as the presence of particular radiological findings. JSON-templates include close-ended questions, such as Yes/No or multichoice questions \cite{Woznicki2024, Mukherjee2023, Song2026}, open-ended questions, for free-text and numerical extraction \cite{Adams2023, Bergomi2024,Song2026}. 
The advantages of these approaches include their high accuracy, cost effectiveness and simple implementation \cite{Adams2023, Mukherjee2023}. Some of these studies highlight the feasibility of using publicly available and locally hosted LLMs to structure radiology reports in a privacy-preserving manner, without additional training or fine-tuning \cite{Adams2023, Woznicki2024}.

A recent study focused on information extraction from CAG reports using a JSON-like template pipeline. The template included including prior stent details, lesion characteristics, anatomical diagnosis, and PCI features (e.g. stent lenght or multivessel PCI). Although they presented great results, their pipeline leverages proprietary LLMs (GPT-4o, Gemini-2.5-Flash and Claude-4.5) that do not guarantee security of the data. Furthermore, they used a small corpus of 150 reports and 100 reports for external validation \cite{Song2026}.


However, several challenges persist in these publications. These include lack of extraction of relevant information, both due to missing details in the templates and ambiguities, or vague wording in the data \cite{Woznicki2024}, and the conclusions are constrained by a limited amount of data \cite{Bergomi2024}. Additionally, semantic understanding also varies across languages and imaging findings, with better results achieved in English corpora compared to German \cite{Woznicki2024}. Furthermore, one important aspect is that numerical data frequently presents inconsistencies in natural language processing tasks \cite{Zhang2020}. Finally, to the best of our knowledge, no studies have investigated structuring radiology reports in Portuguese, and research has primarily focused on modalities such as chest X-rays, with no exploration of coronary angiography.

\section{Problem Formulation}
The task at hand involves extracting the numerical values of FFR and iFR, along with their corresponding measurement locations, from CAG reports.

\subsection{Data}
The dataset consists of 1342 CAG reports from adult patients obtained from the Cardiology Department of Centro Hospitalar Universitário Lisboa Norte (Hospital de Santa Maria), a terciary hospital in Portugal. These reports contain comprehensive documentation of procedures conducted from January 2012 to October 2023 (the distribution of CAG across these years is shown in Figure \ref{year_distribution} in the Appendix), by a team of 15 cardiology specialists. All reports are written in Portuguese.

The dataset comprises extensive information regarding angiography and angioplasty procedures. 
The angiography segment offers an anatomical characterization of coronary arteries and lesions, accompanied by physiological indexes: the FFR and iFR. When the angioplasty is performed, the report includes information regarding the procedure, such as materials utilized, record of any complications, and the evaluation of procedural success. Furthermore, certain reports encompass post-angioplasty physiological assessments and the recommended further treatment.
The groundtruth data consists of a structured data, manually annotated by a team of cardiology residents. The collected features encompass the physiological measurements (FFR and iFR) for the identified lesions.
The dataset comprised FFR and iFR measurements across five coronary vessels. However, each report contained only a small number of measurements (on average, less than 3 fields over 10 possible fields). Thus, our groundtruth was sparse, with a large proportion of missing values. 

An example of such a report is provided in \textit{\textbf{Report 1 - Angiography and Angioplasty}}. The reports can largely differ from the presented example, both in structure and content, depending on the cardiologist. Table \ref{tab:output_example_1}  shows the output of the structured data that we expect to obtain from the report in \textit{\textbf{Report 1 - Angiography and Angioplasty}}, including the measured values of FFR and iFR at each vessel.

\begin{center}
    \noindent\fbox{
        \parbox{0.45\textwidth}{
            \textit{\textbf{Report 1 - Synthetic Report}}
            \textit{CORONARIOGRAFIA. Tronco comum sem lesões. Descendente anterior com lesão longa ligeira a moderada no segmento proximal e médio. Lesão grave, curta, no segmento distal, junto ao apex. Circunflexa dominante com ligeiras irregularidades. Coronária direita de pequeno calibre, não dominante, com lesão crítica no segmento proximal e médio provavelmente culpável pelo síndrome coronário agudo. AVALIAÇÃO FUNCIONAL DA LESÃO LONGA DO SEGMENTO PROXIMAL E MÉDIO DA \textbf{DESCENDENTE ANTERIOR}: Anticoagulação com heparina. Cateter JR 3.5 SH. Fio guia Pressurewire. Adenosia por via endovenosa periférica. \textbf{FFR 0.83} (não significativo). Hemostase do ponto de acesso vascular por compressão. CONCLUSÃO: Doença coronária grave no segmento distal da descendente anterior e coronária direita não-dominante que não justificam intervenção. Lesão longa no segmento proximal e médio da descendente anterior, funcionalmente não significativa. Boa função ventricular esquerda sem alterações segmentares. Terapêutica médica.
    }}
    }
\end{center}


\begin{table}[h]
    \caption{Strutured data output for the report in \textit{\textbf{Report 1 - Angiography and Angioplasty}} }
    \centering
    \begin{tabular}{l|l}
    \toprule
    Field  &  Value  \\ 
    \midrule     
    Tronco Comum FFR         &  \textit{NA} \\
    \textbf{Descendente Anterior FFR} &  \textbf{0.83} \\
    Circunflexa FFR          &  \textit{NA} \\ 
    Coronária Direita FFR           &  \textit{NA} \\ 
    Outras Artérias FFR           &  \textit{NA} \\ 
    Tronco Comum iFR        &  \textit{NA} \\
    Descendente Anterior iFR  &  \textit{NA} \\
    Circunflexa iFR        &  \textit{NA} \\ 
    Coronária Direita iFR          &  \textit{NA} \\ 
    Outras Artérias iFR         &  \textit{NA} \\ 
    \bottomrule
    \end{tabular}
    \label{tab:output_example_1}   
\end{table}

\subsection{Challenges}
The project entails several challenges.
First, it is essential to secure data privacy and security to safeguard patient identities. Only local models safeguard sensitive patient data from exposure to external systems \cite{Kanzawa2024}, limiting the models available for this project.
Furthermore, all the reports are written in portuguese and it is essential to assess the performance of LLMs trained in English and multilingual datasets in our task.
Moreover, medical reports frequently use abbreviations, such as "DA" for "Coronária Descendente Anterior", which might limit the models' capability of extracting the correct values to the correct vessel.

\section{Methodology}
This section describes the methods adopted for the extraction of physiology indexes from CAG reports, comprising the criteria for model selection, the prompting strategies,  benchmarking, an overview the experimental setup and our evaluation pipeline.

\subsection{Model Selection}
Processing real-world clinical data, such as CAG reports, requires rigorous data protection and security measures. To ensure that all sensitive information remained within the institutional environment, only locally deployable models were considered. This requirement restricted model selection to open-source LLMs that allow on-premise inference without transmitting patient data to external servers.

Based on this criteria, three families of foundation models, \textit{Llama 3} (Llama3 8B parameters), \textit{Mistral} (Mistral 7B) and \textit{GPT-OSS} (GPT-OSS 20B), and three medical models, MedGemma (4B), MedLlama (8B) and Med GPT-OSS (20B), were selected for evaluation.
All models were configured with a temperature of 0 to ensure deterministic outputs and no quantization was used in order to obtain the best results possible \cite{huggingface}.

\subsection{Prompting and Output Control Strategies}
To improve extraction performance, prompt engineering techniques were utilized. Both zero-shot (0-S) and few-shot (F-S) prompting approaches were implemented to guide the model toward producing more structured and reliable outputs. The few-shot configuration included three example reports paired with their corresponding extracted values; for simplicity, this setup is referred to as F-S. Furthermore, based on the results of \cite{Adam2025}, we added a third F-S prompt with implausible values for the extracted values, and removed values outside of the known range of FFR/iFR from the extracted values for this approach, to which we refer as F-S Implausible. Additionally, model robustness to prompt variations was
assessed using two similarly designed prompts containing different JSON templates, shown in Appendix Prompts and Templates. 

Furthermore, a post-processing validation step based on regular expressions (RegEx) was applied according to \cite{Adam2025}. This layer verified whether each extracted value was present in the original report, discarding outputs that could not be matched to the original text.

In addition, constrained generation was employed using the Guidance library to enforce output structure during decoding \cite{Guidance}. Constrained generation is a mechanism to enforce structural compatibility of model outputs with downstream requirements and to potentially reduce the computational cost for generation. The mechanism works by sampling next tokens only from the distribution of admissible tokens given a formal grammar, in our case represented by a JSON schema that defines required attributes and data types. If the next token is pre-determined by the grammar, no computation of the next token probabilities is required, reducing the amount of forward-pass computations \cite{Guidance}. However, it has previously been shown that constrained generation can negatively affect reasoning performance \cite{Beurer2024}. The prompts for this approach remained as unchanged as possible compared to the baseline approach (See Appendix Prompts and Templates).

\subsection{Benchmarking}
For comparison against a deterministic baseline, we developed a baseline regular-expression (Baseline RegEx) template tailored to the structure and terminology commonly found in CAG reports. These results served as a reference for evaluating LLM performance.

\subsection{Experimental Setup}
All experiments were performed using the original CAG reports as input to the selected LLMs. 
A structured prompt required the model to complete a predefined JSON-compatible template with FFR and iFR values for each coronary artery. Following extraction, the outputs were parsed and processed by removing unnecessary punctuation and converted into a dataframe to ensure consistency between models and to enable the computation of the evaluation metrics. All preprocessing steps were implemented using deterministic rules to avoid introducing ambiguity into the evaluation pipeline.

\begin{figure}[t]  
    \centering
    \includegraphics[width=0.5\textwidth]{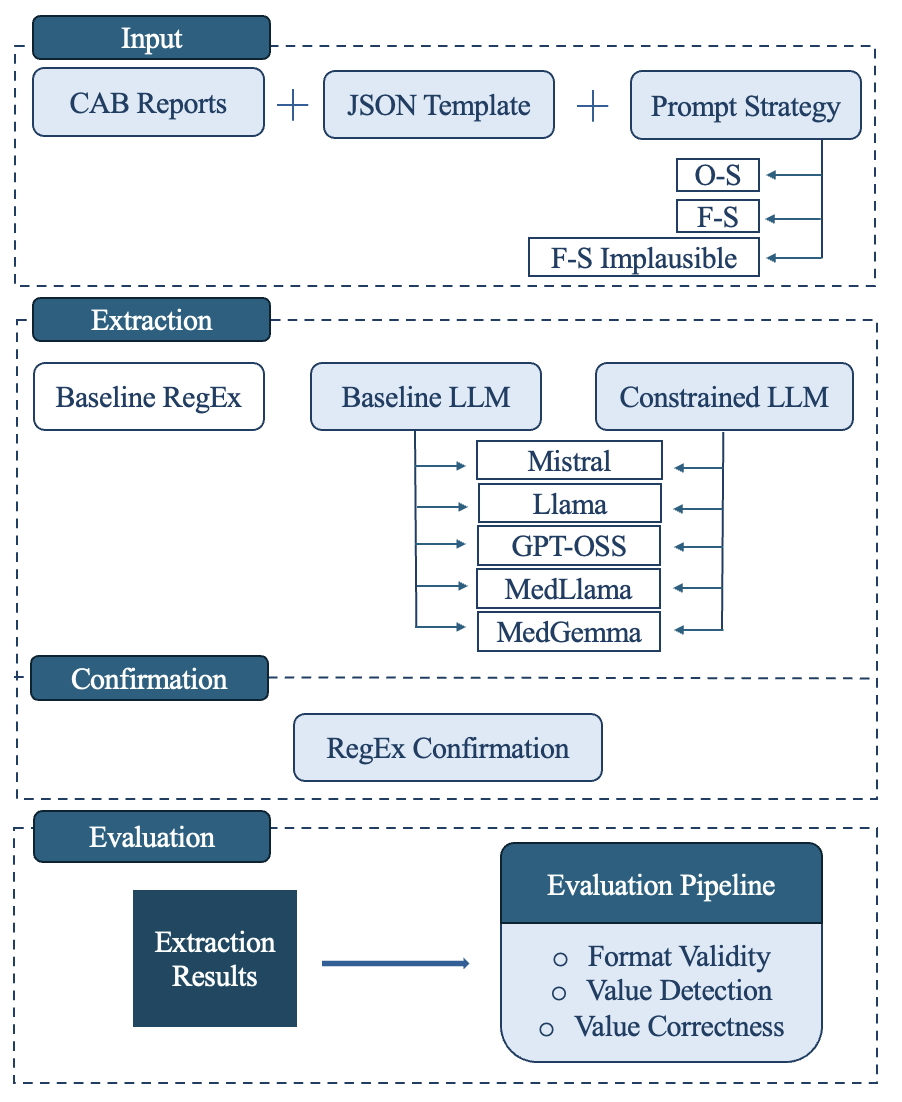}
    \caption{Extraction Pipeline: Extraction was achieved using multiple settings. First, for benchmarking, a Baseline RegEx solution was implemented. The selected LLMs were then applied with each of the three possible prompting techniques. In parallel, the models were run using the library Guidance for constrained generation. Finally, a final RegEx layer was included. The results from all of these setups, with the different ablations, were stored.}
    \label{fig:schema}
\end{figure}

Figure \ref{fig:schema} shows a schema of  our approach: the inputs, which included the reports, JSON template and prompt strategy; the extractors, including the baseline and LLMs, constrained or unconstrained; the confirmation step using RegEx; and the evaluation layer.

We organized our results in four main sections:

\begin{itemize}
\item \textbf{A - Format Validity:} The first step consisted of evaluating the models capability of extracting values in the correct format, respecting a JSON-like template with only numerical values, while comparing constrained and unconstrained generation. Additionally, model robustness to JSON template variations in the prompt was assessed using two carefully designed prompts, in order to give insights on the behaviour of the models for different JSON templates.

\item \textbf{B - Baseline Models:} This experiment aimed at analyzing the possible utility of LLMs in extracting relevant information from CAG reports, comparing with a baseline RegEx methodology. For this, the 3 prompting approaches were utilized: 0-S, F-S and F-S Implausible.

\item \textbf{C - Ablation Studies:} This experiment aimed at evaluating the effect of adding steps to the pipeline, including the usage of a final RegEx step to confirm if the value extracted was indeed in the original report, as well as adding constrained generation using guidance, or both.

\item \textbf{D - Medical Models: } Finally, we aimed at comparing models fine-tuned for the medical domain with general models, as well as to compare the behaviour of two medical models in the previous settings.
\end{itemize}

\subsection{Evaluation Metrics}

Given the characteristics of the extraction task, we designed an evaluation pipeline intended to be comprehensive while reflecting the relative clinical importance of different extraction errors. For this reason, the evaluation extraction was performed in three stages, as is shown in Figure \ref{fig:evaluation_pipeline}:

\begin{figure}[t]  
    \centering
    \includegraphics[width=0.5\textwidth]{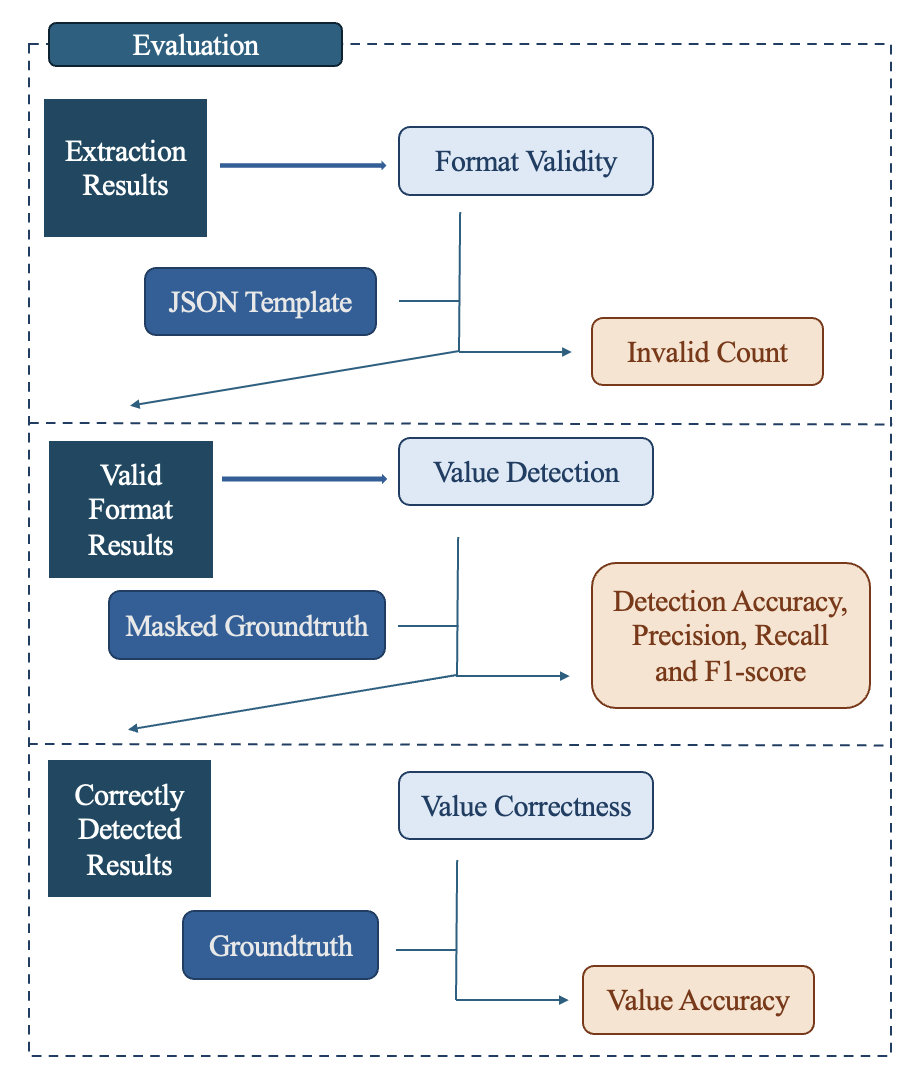}
    \caption{Evaluation Pipeline: The evaluation pipeline is composed of format validity, value detection and value correctness. The metrics used include detection accuracy, precision, recall and F-1 score, and value accuracy.}
    \label{fig:evaluation_pipeline}
\end{figure}

\begin{enumerate}
    \item \textbf{Format Validity:} We first quantified the proportion of outputs that did not comply with the expected JSON-compatible structure or that contained non-numeric entries in numeric fields. These out-of-format responses were considered extraction failures, and were removed from the extraction results. 
    \item \textbf{Value Detection:} For each artery, we assessed whether the model correctly detected the presence/absence of a physiological measurement. For this, a confusion matrix was extracted and accuracy, precision, recall and the F1-score were evaluated. Importantly, false positives and false negatives do not have equivalent implications: incorrectly introducing a value that is not present in the report is worse than failing to extract an existing one. Therefore, in cases where no single model clearly dominates, precision should be weighted more heavily when interpreting performance. 
    \item \textbf{Value Correctness:} When a value was detected in the correct place, accuracy was calculated based on exact correspondence with the groundtruth. Because this task requires precise extraction of clinically meaningful measurements, approximate or close values were considered incorrect.
\end{enumerate}

\section{Results}
The dataset included 1342 reports with a mean length of 197 words (12-667 words). Of these, 692 corresponded to angiography, while 650 also included angioplasty. Regarding the physiology indexes, 10 fields for each report were extracted, corresponding to the measurements of FFR and iFR in 5 different vessels. Out of a total of 13420 fields, 86.4\% of the values were naturally absent. Due to this great class imbalance, it is important to note that a model does not extract any value will have an accuracy of 86.4\%. 
A total 1820 values were measured, 603 corresponding to FFR values and 1217 to iFR values. Table \ref{tab:vessel_measurements} shows the number and percentage of measurements per vessel. 

\begin{table}[htbp]
\centering
\caption{Number and percentage of measurements per vessel and index type (relative to 1,314 reports).}
\label{tab:vessel_measurements}
\begin{tabular}{ll r}
\toprule
Index & Vessel & Count (\%) \\
\midrule
FFR & Left Main               & 16  (1.2\%)  \\
    & Left Anterior Descending & 387 (29.5\%) \\
    & Circumflex              & 91  (6.9\%)  \\
    & Right Coronary         & 95  (7.2\%)  \\
    & Other arteries         & 14  (1.1\%)  \\
iFR & Left Main               & 32  (2.4\%)  \\
    & Left Anterior Descending & 720 (54.8\%) \\
    & Circumflex              & 188 (14.3\%) \\
    & Right Coronary         & 224 (17.0\%) \\
    & Other arteries         & 54  (4.1\%)  \\
\bottomrule
\end{tabular}
\end{table}

\subsection{Format Validity}
Format validity was assessed across all models and configurations. MedLlama (0-S, F-S, F-S Implausible) was unable to extract any values in the correct format with any prompt strategy, while MedGemma produced a small number of outputs that did not conform to the expected template. However, this limitation was resolved by applying constrained generation with Guidance, as is shown in Table \ref{tab:outofformat}. The remaining models did not have any out-of-format results regarding of the usage of constrained generation.

\begin{table}[h]
\centering
\caption{Number of out-of-format extractions per model}
\begin{tabular}{lllll}
\toprule
Model & Baseline & Constrained \\
\midrule
GPT-OSS 0/F-S/F-S I & 0  &  0 \\
Mistral 0/F-S/F-S I & 0 &  0 \\
Llama 0/F-S/F-S I &  0 &  0 \\
MedGemma 0-S & 2 &  0 \\
MedGemma F-S & 8 &  0  \\
MedGemma F-S I & 2  &  0  \\
Medllama 0/F-S/F-S I & 1342  &  0 \\
\bottomrule
\end{tabular}
\label{tab:outofformat}
\begin{tablenotes}
    \centering
    \item 0-S - Zero-shot; F-S - Few-Shot;  F-S I - Few-Shot Implausible.
    \end{tablenotes}
\end{table}

Regarding Detection and Value Correctness, 3 main experiments were performed, and the results are shown in the next sections.

\subsection{Baseline Models and Prompting}
Table \ref{tab:results_baseline_strategy} shows the results of baseline RegEx and general models (without constrained generation or RegEx step) using the three proposed prompting strategies: 0-S, 1-S and 1-S Implausible. 
The three models outperform the RegEx baseline, while showing only small performance differences among themselves. Llama and GPT-OSS in the zero-shot setting achieve the best performance on the extraction metrics, whereas Mistral has the highest extracted value accuracy.
Prompting strategies show only minor differences in performance. In particular, adding few-shot examples did not consistently improve results and neither did the inclusion of implausible values in the prompt compared to few-shot prompting.

\begin{table*}[htbp]
\centering 
\caption{Baseline models with three prompt strategies}
\begin{tabular}{l|llll|l}
\toprule
  & Extraction  &  &  & & Value \\
Model & Accuracy & Precision & Recall & F1 & Accuracy\\
\midrule
Baseline RegEx & 0.943 & 0.832 &  0.727  &  0.776 & 0.888\\
\midrule
Mistral 0-S  & \underline{0.966} &  0.867   & 0.881  & 0.874  & 0.939  \\
Mistral F-S   & 0.965 &  0.869   & 0.871  &  0.870  & \textbf{0.943} \\
Mistral F-S Implausible values & 0.965 & 0.868    &  0.875 &  0.871  & \textbf{0.943} \\
Llama 0-S  & \textbf{0.967} &   \textbf{0.872}  & \textbf{0.885} & \textbf{0.878} &  \underline{0.941} \\
Llama F-S  &  0.965   &  0.865 &  0.877  & 0.871 & 0.939 \\
Llama F-S Implausible values &  0.965 &  0.870   &  0.872 &  0.871  & 0.938  \\
GPT-OSS 0-S   & \underline{0.966} & \underline{0.871} & \underline{0.884}  &  \underline{0.877} &  0.939 \\
GPT-OSS F-S  & 0.965  &  0.865   & 0.877 &  0.871  &  0.939 \\
GPT-OSS F-S Implausible values  & 0.964 &  0.866   & 0.873  &  0.869  &  \underline{0.941} \\
\bottomrule
\end{tabular}
\label{tab:results_baseline_strategy}
\begin{tablenotes}
    \centering
    \item  0-S - zero-shot; F-S - few-shot; Implausible - adding implausible values to the few-shot prompt; RegEx- Regular Expressions.
    \end{tablenotes}
\end{table*}



\begin{figure}[t]  
    \centering
    \includegraphics[width=0.47\textwidth]{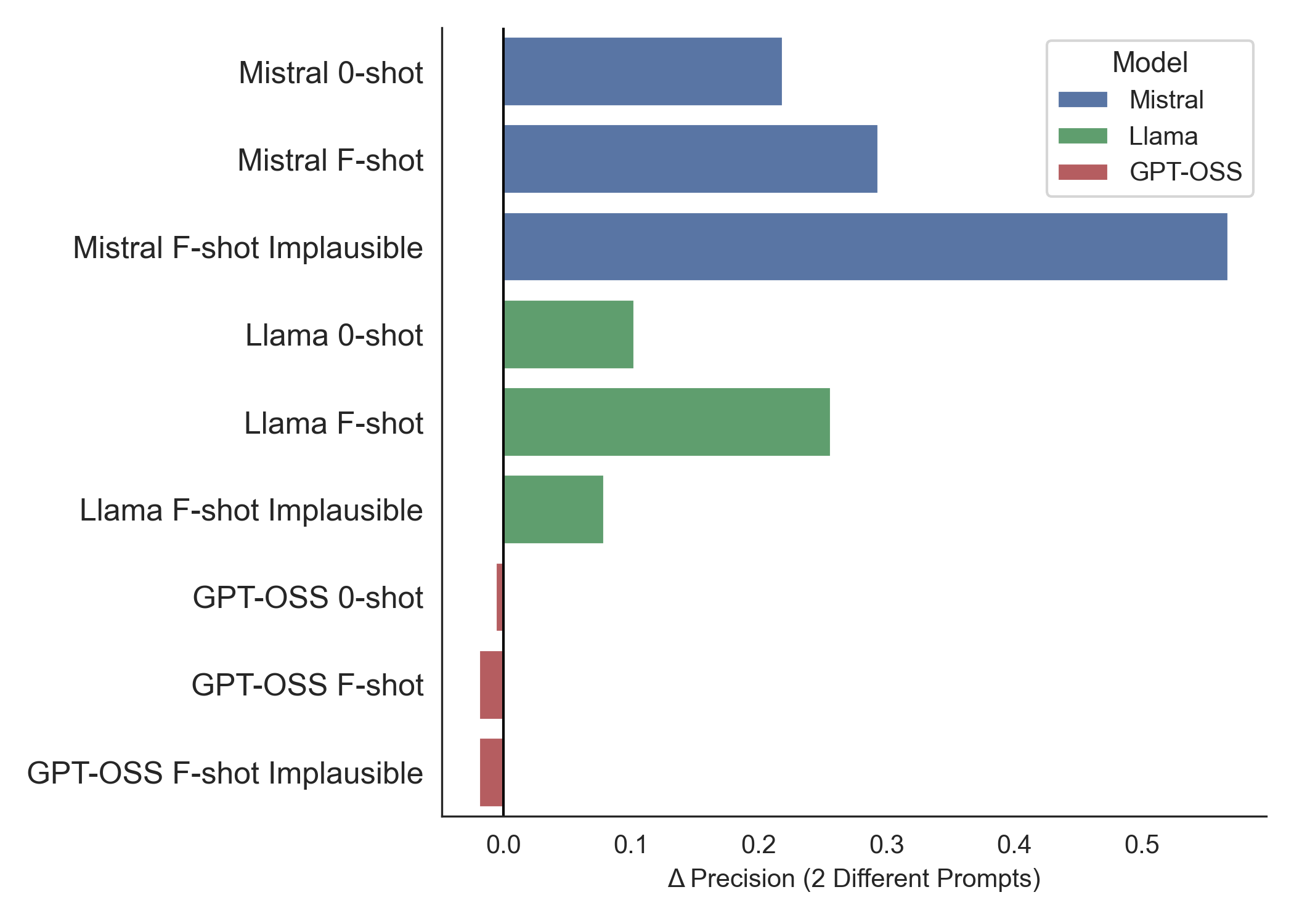}
    \caption{Comparison of model robustness to changes in prompt: variation in precision}
    \label{fig:metric_prompt_robustness}
\end{figure}

\begin{figure}[t]  
    \centering
    \includegraphics[width=0.47\textwidth]{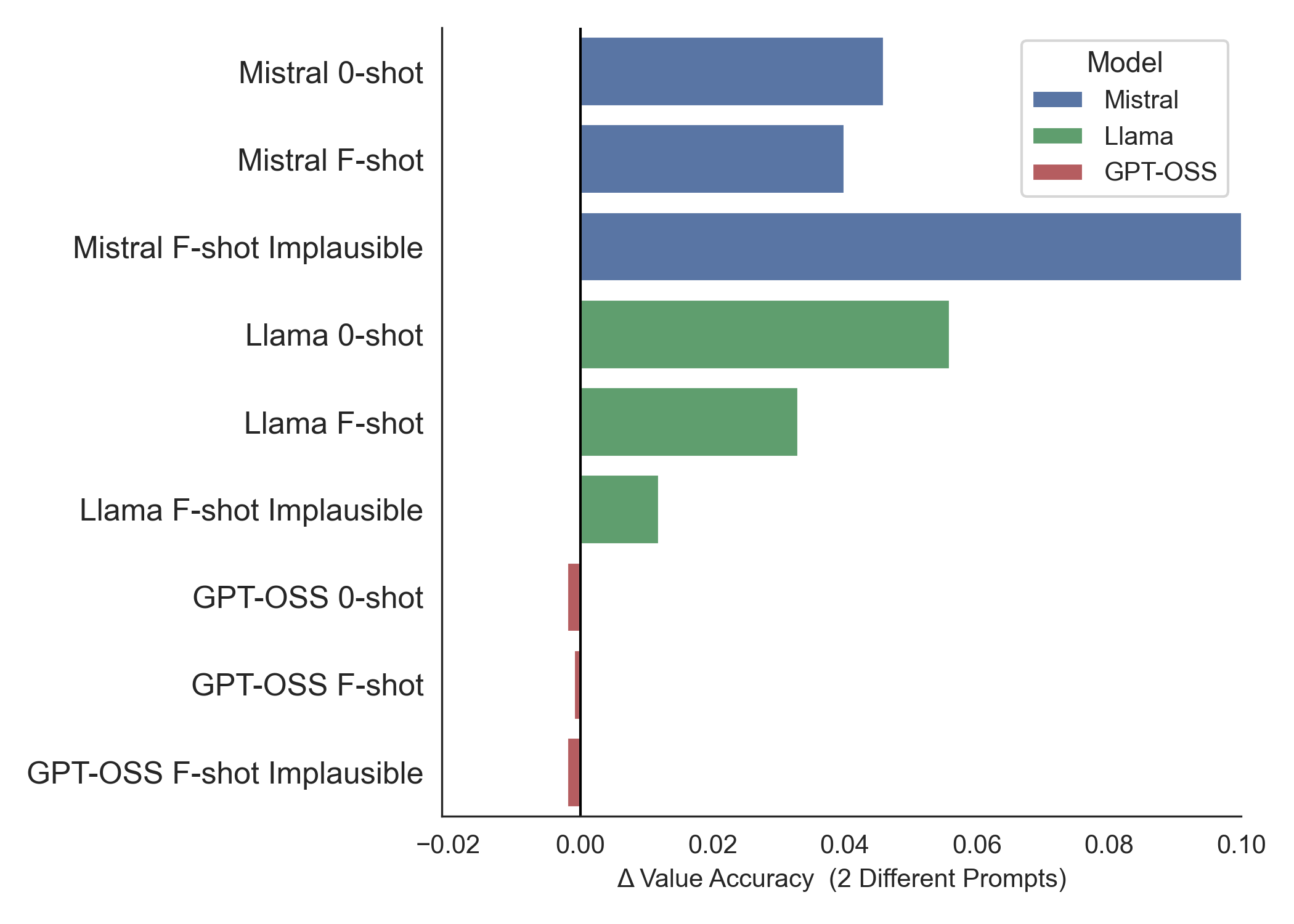}
    \caption{Comparison of model robustness to changes in prompt: variation in value accuracy}
    \label{fig:value_acc_prompt_robustness}
\end{figure}

Two different prompts (see Appendix) differing only in the structure of the JSON template were used to test robustness of the models to prompt variations. Figures \ref{fig:metric_prompt_robustness} and \ref{fig:value_acc_prompt_robustness} present the variation in detection precision and value accuracy across the two prompts. GPT-OSS showed the highest robustness to these changes, while Mistral, especially with the few-shot Implausible prompt, was very sensitive to these changes.

\subsection{Ablation Studies}
The ablation study results for adding the RegEX step and the constraint generation are shown below.

Figure \ref{fig:adding_regex} shows the results of adding a RegEx step over all metrics are very limited, and, in fact, a small decrease in the detection metrics can be seen. The slight decrease in recall and, consequently, F1-score was expected, as we are indeed excluding values that had been previously extracted. In the case any value was extracted from a measurement that had been made, then by removing it, we are indeed decreasing the extraction recall. Nevertheless, this slight decrease is mostly accompanied by an equivalently small increase in the accuracy of the extracted value, since only values not present in the report are excluded. 

\begin{figure*}[htbp]  
    \centering
    \includegraphics[width=\textwidth]{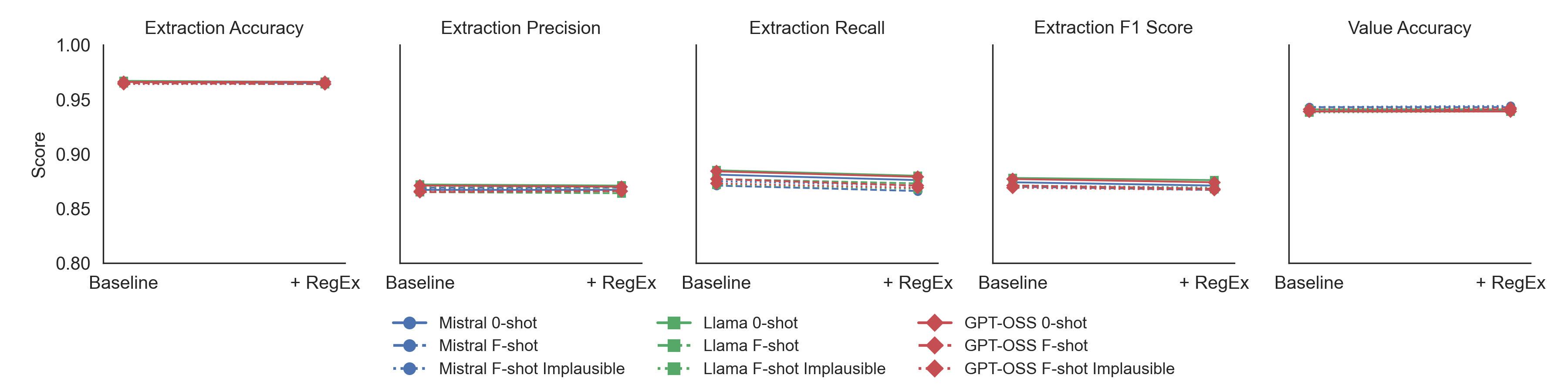}
    \caption{Comparison between baseline and adding confirmation step using RegEx. Detection metrics (Accuracy, Precision, Recall, F1) and value extraction accuracy are shown.}
    \label{fig:adding_regex}
\end{figure*}

\begin{figure}[h]  
    \centering
    \includegraphics[width=0.47\textwidth]{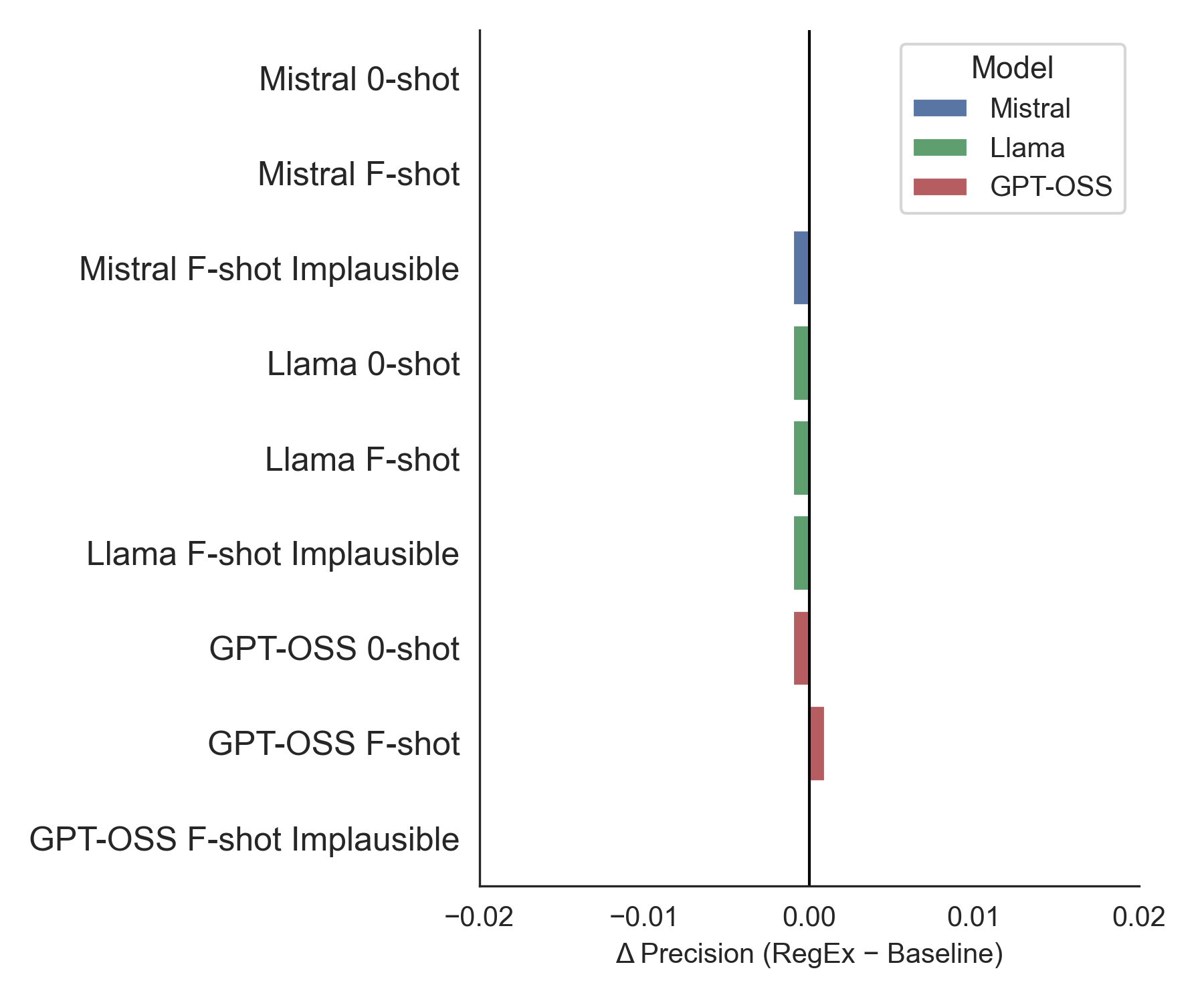}
    \caption{Difference in precision by adding confirmation step using RegEx.}
    \label{fig:precision_regex_delta_horizontal}
\end{figure}

Taking into account the clinical context, the most critical type of error in this context was to reduce false extractions, corresponding to the extractions of values where no measurement had been made. For that reason, a RegEx confirmation layer was tested for precision. The results are presented in Figure \ref{fig:precision_regex_delta_horizontal}. 

Unexpectedely, we observe the oposite effect for this layer, with a slight decrease in precision for some models. This behaviour is explained by the removal from the extraction of values corresponding to typographical errors in the report, that thus RegEx did not identify. The 5 types of identified typos are shown in Table \ref{tab:number_format_corrections}. These examples showed LLMs are able to identify measurements and correct the typos, while deterministic approaches such as RegEx may fail. Furthermore, this shows the importance of being able to identify where the LLM is extracting the values from, as we could have easily understood that the extracted value is a corrected version of the value in the text (and thus correctly annoted in our groundtruth).

\begin{table}[ht]
\centering
\caption{Examples of typographical errors identified in the corpus and the corresponding LLM extraction. }
\begin{tabular}{lll}
\hline
\textbf{Type of typographical error} & \textbf{Report} & \textbf{Extracted} \\
\hline
Missing "0" & ,93      & 0.93 \\
Missing Comma &  089      & 0.89 \\
Extra Space  & 0, 75    & 0.75 \\
Wrong Comma Placing & ,099     & 0.099 \\
Repeated "0" and Punctuation & 0,0.93   & 0.93 \\
\hline
\end{tabular}
\label{tab:number_format_corrections}
\end{table}

As shown in Table~\ref{tab:outofformat}, adding constrained generation was crucial to obtain extraction values that conformed with the required format. 
Table \ref{tab:presence_absence_results} in the Appendix shows the results of constrained generation using Guidance. The general tendency is a decrease in all metrics using guidance, especially for extraction recall and F1-score consequently. Nevertheless, this approach allows for the usage of models that do not comply with the JSON template in an unconstrained scenario.

\begin{figure*}[t]  
    \centering
    \includegraphics[width=\textwidth]{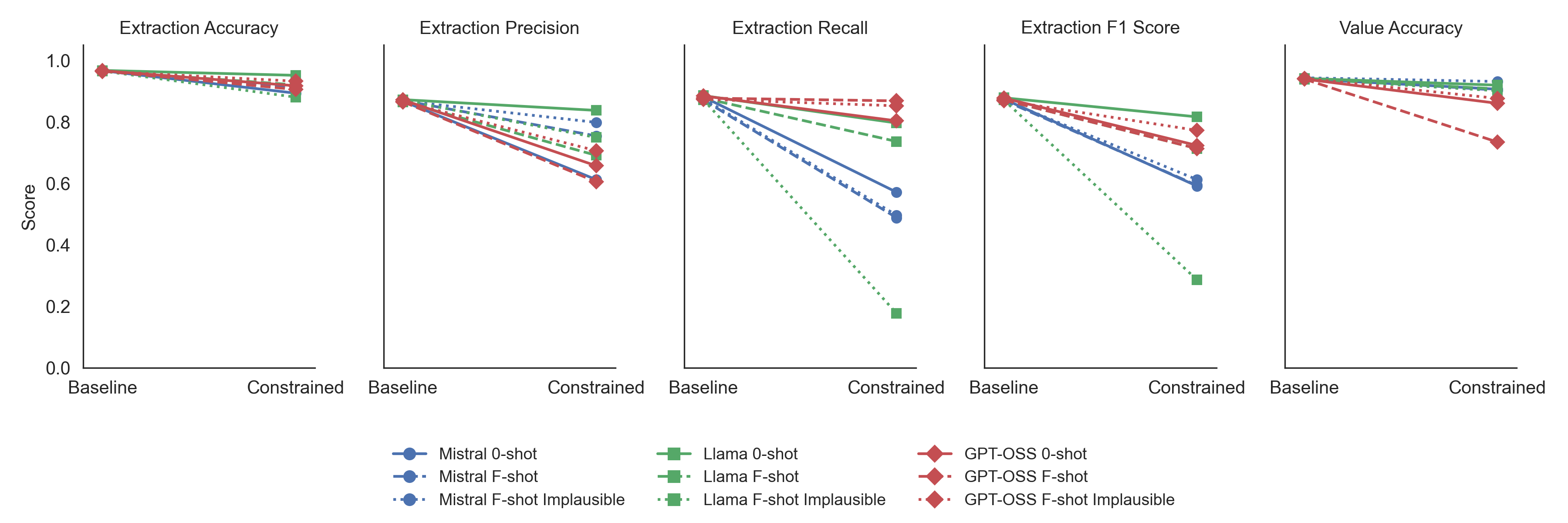}
    \caption{Comparison between prompting strategies for the baseline models: zero-shot (0-S), a few-shot (F-S) and few-shot with examples using Implausible values for the context (1-S Implausible). Detection metrics (Accuracy, Precision, Recall, F1) and value extraction accuracy are shown.}
    \label{fig:adding_guidance}
\end{figure*}

A complete ablation study analysis was performed for GPT-OSS, which presented good performance and more stability to JSON schema changes, and is shown in Table \ref{tab:presence_absence_results}. Figures \ref{fig:gpt_ablation_0-S}, \ref{fig:gpt_ablation_F-S} and \ref{fig:gpt_ablation_F-S_implausible} illustrate the ablation results for GPT-OSS. While the RegEx layer as little impact on the model performance, adding contrained generation lead to a decrease in the metrics.

\begin{table*}[p]
\centering
\caption{GPT-OSS Ablation Study}
\renewcommand{\arraystretch}{0.9} 
\begin{tabular}{l|llll|l}
\toprule
& Extraction  &  &  & & Value \\
Model  & Accuracy & Precision & Recall & F1 & Accuracy \\
\midrule
Baseline RegEx & 0.943 & 0.832 &  0.727  &  0.776 & 0.888\\
\midrule
GPT-OSS 0-S   & \textbf{0.966}& \textbf{0.871} & \textbf{0.884}  &  \textbf{0.877}  &  0.939 \\
GPT-OSS F-S  & \underline{0.965}  &  0.865   & 0.877 &  0.871  &  0.939 \\
GPT-OSS F-S Implausible values  & 0.964 &  0.866   & 0.873  &  0.869  &  \underline{0.941}  \\
\midrule 
GPT-OSS 0-S + RegEx  & \textbf{0.966} &  \underline{0.870}  & \underline{0.879 } & \underline{0.874 }  & 0.939  \\
GPT-OSS F-S + RegEx  & 0.964 &  0.866   & 0.871  & 0.868   & \underline{0.941}  \\
GPT-OSS F-S Implausible values + RegEx  &  0.964 &  0.866   & 0.869  &     0.867 & \textbf{0.942} \\
\midrule
GPT-OSS 0-S Constrained & 0.917  &   0.657   &   0.803  &   0.723 & 0.860  \\
GPT-OSS F-S Constrained & 0.905  &   0.605   &   0.868  &   0.713 & 0.734  \\
GPT-OSS F-S Implausible values  Constrained & 0.932  &   0.706   &   0.852  &   0.772 & 0.876 \\
\midrule
GPT-OSS 0-S  Constrained + RegEx & 0.948  &   0.858  &   0.739  &   0.794 & 0.930  \\
GPT-OSS F-S  Constrained + RegEx & 0.936  &   0.805   &   0.695  &   0.746 & 0.912 \\
GPT-OSS F-S Implausible values  Constrained + RegEx & 0.950 &   0.820   &   0.810  &   0.815 & 0.918 \\
\bottomrule
\end{tabular}
\label{tab:presence_absence_results}
\begin{tablenotes}
    \centering
    \item  0-S - zero-shot; Constrained - Constrained Generation with Guidance; F-S - few-shot; Implausible - adding implausible values to the few-shot prompt; RegEx- Regular Expressions. Settings with no extraction where not included in this table. The best results for each metric are shown in bold, while underline results represent the second best.
    \end{tablenotes}
\end{table*}

\subsection{Medical Models}
As mentioned previously and shown in Table \ref{tab:outofformat}, MedLlama was uncapable of extracting in-format values without constrained generation, hindering its usage in this context. Similarly, in a few cases, MedGemma produced out-of-format examples. 

Furhtermore, MedLlama shows overall weaker performance across most metrics, while the remaining best model per family performs similarly.


Table \ref{tab:presence_absence_results_medical} shows the results for the experiments with the medical models. Results for unconstrained MedLlama were not possible to obtain as they were out of format. Likewise, the results for constrained zero-shot followed by a RegEx step resulting in 0 extracted values, and the observed accuracy is due to the presence of 86.4\% of missing values in the groundtruth, corresponding to values that were not measured. 

MedGemma outperformed MedLlama in all settings and obtained results comparable to those of non-medical models. The best results were obtained using MedGemma with zero-shot prompting and RegEx confirmation.

\begin{table*}[htbp]
\centering
\caption{FFR/iFR values extraction results using Medical LLMs}
\renewcommand{\arraystretch}{0.9} 
\begin{tabular}{l|llll|l}
\toprule
& Extraction  &  &  & & Value \\
Model  & Accuracy & Precision & Recall & F1 & Accuracy \\
\midrule
Baseline RegEx & 0.943 & 0.832 &  0.727  &  0.776 & 0.888\\
\midrule
MedGemma 0-S  & \textbf{0.967}  &  \textbf{0.874}   & \textbf{0.885}  &  \textbf{0.880}  &  \underline{0.940}  \\
MedGemma F-S  & 0.965  & 0.865 & 0.877  & 0.871   & 0.939  \\
MedGemma F-S Implausible  & 0.965 & \underline{0.870}   & 0.872  & 0.871 & 0.938   \\
MedGemma 0-S + RegEx  & \underline{0.966} &  \textbf{0.874} & \underline{0.880}  & \underline{0.877} & \textbf{0.941}  \\
MedGemma F-S + RegEx  & 0.964 & 0.864 &  0.873 &  0.868 &  \underline{0.940} \\
MedGemma F-S Implausible + Regex  & 0.964 & 0.869  & 0.868  &  0.869  & 0.939  \\
\midrule
Constrained Generation & & & & & \\
\midrule
MedLlama 0-S Constrained & 0.864 & 0.500 & 0.001 & 0.001 & 0.00 \\
MedLlama F-S Constrained &  0.780  & 0.198  &  0.202 & 0.200 & 0.620 \\
MedLlama F-S Implausible Constrained & 0.880 & 0.749 & 0.177 & 0.286 & 0.901 \\
MedGemma 0-S Constrained & 0.871  &   0.526   &   0.509 &   0.517 & 0.834   \\
MedGemma F-S Constrained &  0.856  &   0.476   &  0.599   &  0.531 & 0.838  \\
MedGemma F-S Implausible Constrained & 0.900 & 0.649 & 0.573 & 0.608 & 0.874 \\
MedLlama 0-S  Constrained + RegEx & 0.864 & 0.000 & 0.000 & 0.000 & N/A \\
MedLlama F-S  Constrained + RegEx &  0.842 & 0.333  &  0.165 & 0.221  & 0.751 \\
MedLlama F-S Implausible Constrained + RegEx & 0.880 & 0.754 & 0.176 & 0.286 & 0.903 \\
MedGemma 0-S  Constrained + RegEx & 0.877  &   0.552 &   0.499 &   0.524  & 0.847  \\
MedGemma F-S  Constrained + RegEx &  0.871  &   0.523 &  0.585 & 0.552 & 0.852  \\
MedGemma F-S Implausible  Constrained + RegEx & 0.903 & 0.669 & 0.566 & 0.614 & 0.881 \\
\bottomrule
\end{tabular}
\label{tab:presence_absence_results_medical}
\begin{tablenotes}
    \centering
    \item  0-S - zero-shot; Constrained - Constrained Generation with Guidance; F-S - few-shot; Implausible - adding implausible values to the few-shot prompt; NA - Not Applicable; RegEx- Regular Expressions. The best results for each metric are shown in bold, while underline results represent the second best.
    \end{tablenotes}
\end{table*}


The complete results of all experiments, including the accuracy, precision, recall and F1 score for the detection of FFR or iFR values and the accuracy of the extracted values, are shown in Table \ref{tab:presence_absence_results} in the Appendix.

\section{Limitations}
First, the dataset may harbor inherited biases. It comprised reports from 15 cardiology specialists, all sourced from a single center. The format and style of these reports may vary from those of other hospitals, thereby restricting the model's generalizability to different scenarios. Furthermore, our center status as a tertiary and University Hospital may introduce bias and influence the heterogeneity of the cases. 

A small number of models were used in this evaluation and more powerful models, with a larger number of parameters, could outperform the results obtained. Nevertheless, our approach was limited to models we believe might be used in hospital premises, as they do not rely on data-center scale compute capacity. Furthermore, due to the reasons exposed, no portuguese models were considered, and this study was limited to the application of English-based or multilingual based models.

Finally, CAG reports contain additional clinically relevant information that could be valuable to extract. Due to the large dataset size and the time-intensive nature of manual groundtruth annotation, we restricted this study to a subset of physiological measurements, due to their importance in treatment decision. The extraction and evaluation of a broader set of report elements was therefore beyond the scope of this work and remains an important direction for future research.

\section{Conclusions}
This study demonstrates the potential of LLMs for extracting structured information from Portuguese CAG reports, consistently outperforming a RegEx baseline, with the exception of MedLlama. The best overall results were obtained with Llama 0-shot. Interestingly, all models performed similarly independently of their size. As expected, changes in prompt had an impact in the model performance, particularly for smaller models. Nevertheless, GPT-OSS showed more robustness to small changes in the prompts. Adding a RegEx layer showed little to no improvement, while using constrained generation decreases performance, while allowing the usage of models that do not produce in format results. MedGemma showed comparable results to non-medical models.

\section{References}
\bibliographystyle{unsrt}
\bibliography{all_bib}

\appendix

\begin{figure}[htbp]  
    \centering
    \includegraphics[width=0.3\textwidth]{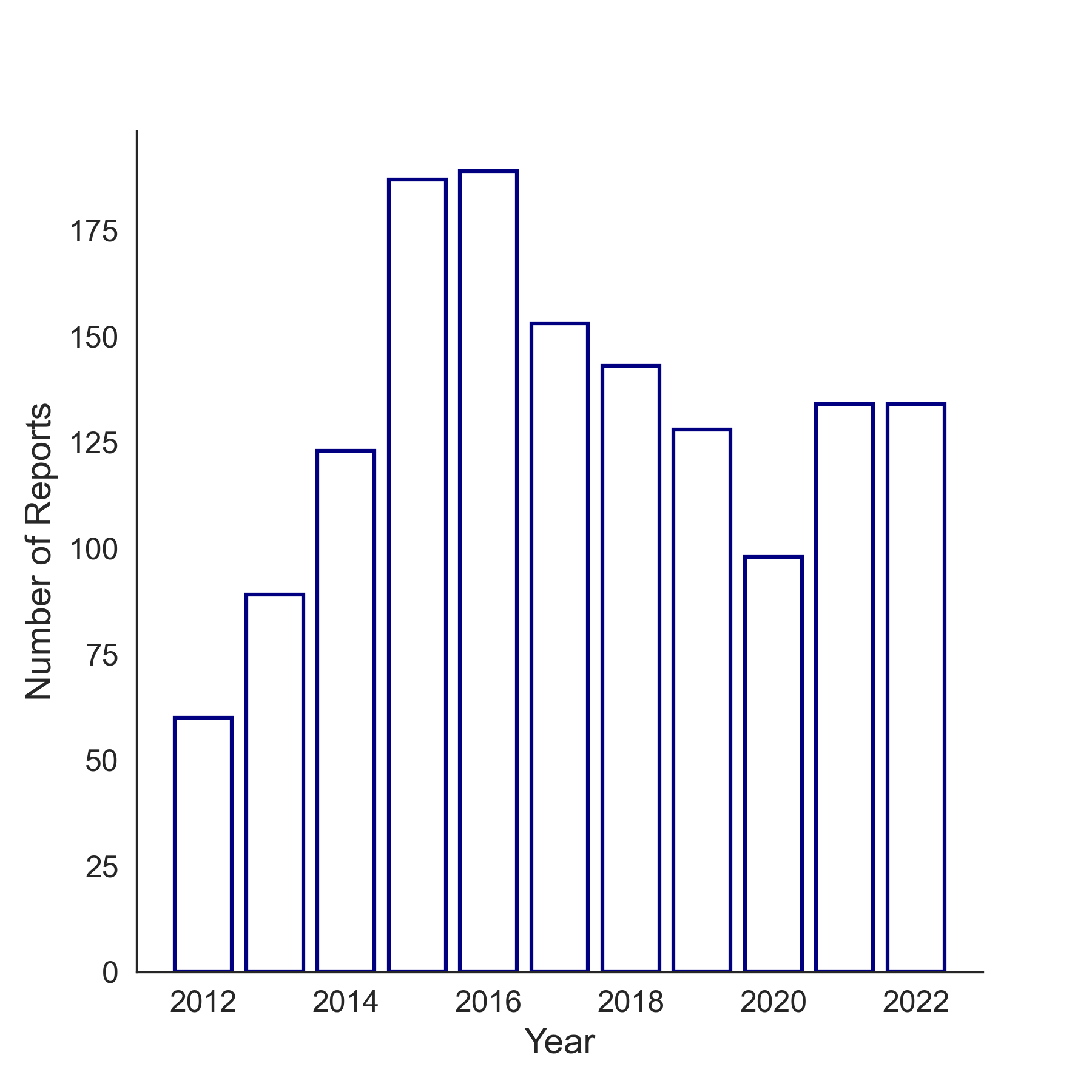}
    \centering
    \caption{Distribution of the CAG procedures from Hospital de Santa Maria per year.}
    \label{year_distribution}
\end{figure}

\begin{center}
\begin{minipage}{0.49\textwidth}
\textit{\textbf{Original Zero-Shot Prompt}}
\begin{Verbatim}[breaklines=true, breaksymbolleft={}, breaksymbolright={}]

És um especialista em relatórios de angiografia/coronariografia e angioplastia. O relatório inclui informação relativa aos indices de fisiologia: fractional flow reserve ou FFR e instant wave-free ratio ou iFR. Retira do texto os valores corretos de FFR e iFR e completa o seguinte template em JSON com os valores corretos para cada artéria (quando o valor não está presente no relatório, mantém "null"). Template:
    {
    "Tronco_Comum_FFR": null,
    "Tronco_Comum_iFR": null,
    "Descendente_Anterior_FFR": null,    
    "Descendente_Anterior_iFR": null, 
    "Circunflexa_FFR": null,
    "Circunflexa_iFR": null,
    "Coronária_Direita_FFR": null,
    "Coronária_Direita_iFR": null,
    "Outras_artérias_FFR": null,
    "Outras_artérias_iFR": null
    }
Relatório: |<REPORT>|
Responde apenas com o template preenchido, sem incluir mais texto na tua resposta, e sem justificar. A tua resposta deve ser um JSON válido.
Para cada valor de FFR ou iFR do template, deves apenas colocar um número decimal. Nenhum outro tipo de informação deve ser preenchida, como palavras ou frases. Nunca coloques um valor que não está presente no relatório. Em caso de dúvida, preenche com null. Quando há mais de uma medição de FFR ou iFR para a mesma artéria, inclui o valor mais baixo.

\end{Verbatim}
\end{minipage}
\end{center}

\begin{center}
\begin{minipage}{0.49\textwidth}
\textit{\textbf{Constrained Zero-Shot Prompt}}
\begin{Verbatim}[breaklines=true, breaksymbolleft={}, breaksymbolright={}]

És um especialista em relatórios de angiografia/coronariografia e angioplastia.O relatório inclui informação relativa aos indices de fisiologia: fractional flow reserve ou FFR e instant wave-free ratio ou iFR. Dá como output um JSON válido. Usa estas descrições como referência: [Field descriptions from JSON Output Schema]

\end{Verbatim}
\end{minipage}
\end{center}

\begin{center}
\begin{minipage}{0.45\textwidth}
\textit{\textbf{JSON Schema for Constrained Generation}}

\begin{Verbatim}[breaklines=true, breaksymbolleft={}, breaksymbolright={}]

{
  "schema": "http://json-schema.org/draft-07
  /schema#",
  "type": "object",
  "properties": {
    "Tronco_Comum_FFR": {
      "type": ["number", "null"],
      "description": "Extrai o valor de FFR no Tronco Comum. Extrair apenas o valor decimal. Nunca incluir valores não presentes no relatório. Em caso de dúvida, preencher com null. Quando há mais de uma medição, incluir o valor mais baixo."
    },

    [...]

    "Tronco_Comum_iFR": {
      "type": ["number", "null"],
      "description": "Extrai o valor de iFR no Tronco Comum. Extrair apenas o valor decimal. Nunca incluir valores não presentes no relatório. Em caso de dúvida, preencher com null. Quando há mais de uma medição, incluir o valor mais baixo."
    },

    [...]
  },
}
\end{Verbatim}
\end{minipage}
\end{center}

\begin{center}
\begin{minipage}{0.45\textwidth}
\textit{\textbf{Original Zero-Shot Prompt - Second JSON Template}}
\begin{Verbatim}[breaklines=true, breaksymbolleft={}, breaksymbolright={}]
[...]
Template:
    {
  "Tronco Comum": {
    "FFR": null,
    "iFR": null
  },
  "Descendente Anterior": {
    "FFR": null,
    "iFR": null
  },
  "Circunflexa": {
    "FFR": null,
    "iFR": null
  },
  "Coronária Direita": {
    "FFR": null,
    "iFR": null
  },
  "Outras artérias": {
    "FFR": null,
    "iFR": null
  }
}
[...]
\end{Verbatim}
\end{minipage}
\end{center}

\begin{figure}[htbp]  
    \centering
    \includegraphics[width=0.45\textwidth]{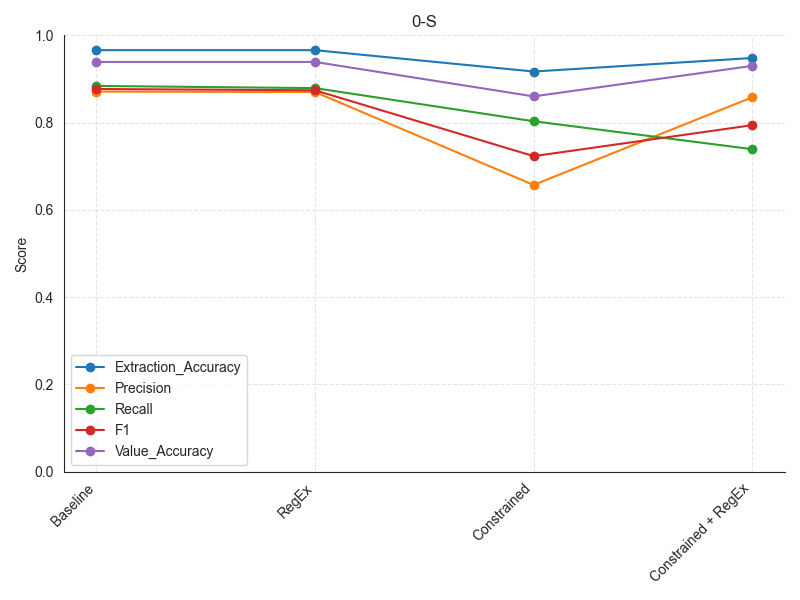}
    \caption{Results for the ablation study of zero-shot GPT-OSS. 0-S - zero-shot.}
    \label{fig:gpt_ablation_0-S}
\end{figure}

\begin{figure}[htbp]  
    \centering
    \includegraphics[width=0.45\textwidth]{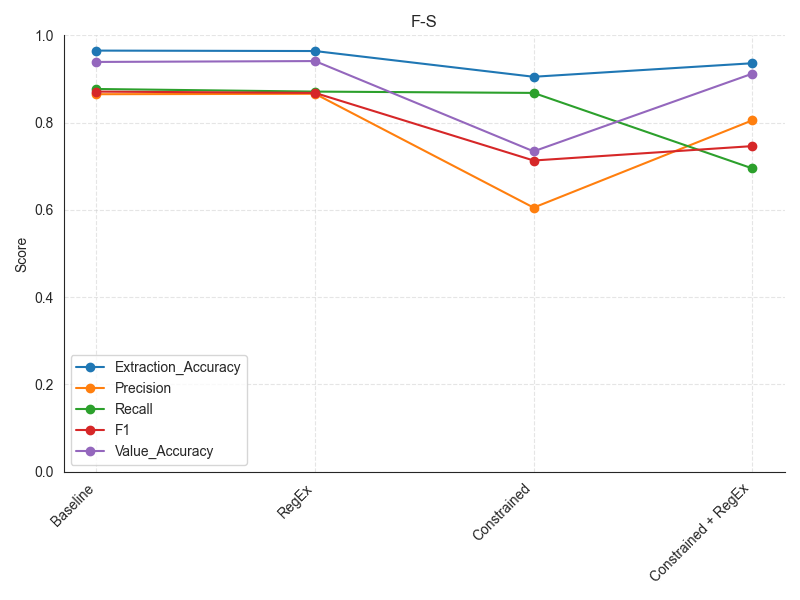}
    \caption{Results for the ablation study of few-shot GPT-OSS. F-S - few-shot.}
    \label{fig:gpt_ablation_F-S}
\end{figure}

\begin{figure}[htbp]  
    \centering
    \includegraphics[width=0.45\textwidth]{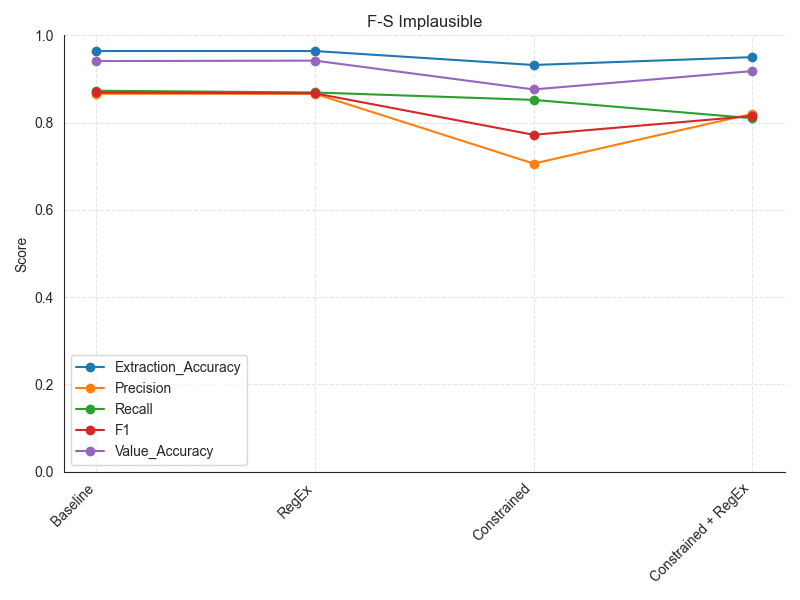}
    \caption{Results for the ablation study of few-shot Implausible GPT-OSS. F-S - few-shot.}
    \label{fig:gpt_ablation_F-S_implausible}
\end{figure}

\begin{table*}[p]
\centering
\caption{FFR/iFR values extraction results}
\footnotesize 
\renewcommand{\arraystretch}{0.9} 
\begin{tabular}{l|llll|l}
\toprule
& Extraction  &  &  & & Value \\
Model  & Accuracy & Precision & Recall & F1 & Accuracy \\
\midrule
Baseline RegEx & 0.943 & 0.832 &  0.727  &  0.776 & 0.888\\
\midrule
Mistral 0-S  & 0.966 &  0.867   & 0.881  & 0.874  & 0.939  \\
Mistral F-S   & 0.965 &  0.869   & 0.871  &  0.870  & 0.943 \\
Mistral F-S Implausible values & 0.965 & 0.868    &  0.875 &  0.871  & 0.943  \\
Llama 0-S  & 0.967 &   0.872  &  0.885 &  0.878  &  0.941 \\
Llama F-S  &  0.965   &  0.865 &  0.877  & 0.871 & 0.939 \\
Llama F-S Implausible values &  0.965 &  0.870   &  0.872 &  0.871  & 0.938  \\
GPT-OSS 0-S   & 0.966 & 0.871 & 0.884  &  0.877  &  0.939 \\
GPT-OSS F-S  & 0.965  &  0.865   & 0.877 &  0.871  &  0.939 \\
GPT-OSS F-S Implausible values  & 0.964 &  0.866   & 0.873  &  0.869  &  0.941 \\
MedGemma 0-S  & 0.967  &  0.874   & 0.885  &  0.880  &  0.940 \\
MedGemma F-S  & 0.965  & 0.865 & 0.877  & 0.871   & 0.939  \\
MedGemma F-S Implausible  & 0.965 & 0.870   & 0.872  & 0.871 & 0.938   \\

\midrule
Confirmation with RegEx  & & & & & \\
\midrule
Mistral 0-S   + RegEx  & 0.965 &   0.867  & 0.876  &  0.871 & 0.940  \\ 
Mistral F-S  + RegEx  &   0.964 & 0.869  & 0.866  & 0.867   & 0.943  \\ 
Mistral F-S Implausible values + RegEx  & 0.964 & 0.867    &  0.871 &  0.869  &  0.944  \\
Llama 0-S  + RegEx  & 0.966 &  0.871  &  0.880 &  0.876  &  0.941 \\ 
Llama F-S  + RegEx  &  0.964  & 0.864 &  0.873 &  0.868  & 0.940  \\  
Llama F-S Implausible values + RegEx   &  0.964 &  0.869   &  0.868 &  0.869  & 0.939  \\
GPT-OSS 0-S + RegEx  & 0.966 &  0.870   & 0.879  & 0.874   & 0.939  \\
GPT-OSS F-S + RegEx  & 0.964 &  0.866   & 0.871  & 0.868   & 0.941  \\
GPT-OSS F-S Implausible values + RegEx  &  0.964 &  0.866   & 0.869  &     0.867 & 0.942 \\
MedGemma 0-S + RegEx  & 0.966 &  0.874 & 0.880  & 0.877 & 0.941  \\
MedGemma F-S + RegEx  & 0.964 & 0.864 &  0.873 &  0.868 &  0.940 \\
MedGemma F-S Implausible + Regex  & 0.964 & 0.869  & 0.868  &  0.869  & 0.939  \\
\midrule
Constrained Generation & & & & & \\
\midrule
Mistral 0-S  Constrained & 0.893  &   0.612   &   0.571 &   0.591  & 0.906  \\
Mistral F-S Constrained &  0.909  &   0.754   &  0.488   &   0.593 & 0.903   \\
Mistral F-S Implausible values Constrained & 0.915  &   0.798   &   0.496  &   0.612 & 0.931  \\
Llama 0-S Constrained &  0.951  & 0.837  &  0.796 & 0.816 & 0.919 \\
Llama F-S Constrained &  0.920  & 0.691  &  0.736 & 0.713 & 0.907 \\
Llama F-S Implausible values Constrained & 0.880  &   0.749   &   0.177  &   0.286 & 0.901 \\
GPT-OSS 0-S Constrained & 0.917  &   0.657   &   0.803  &   0.723 & 0.860  \\
GPT-OSS F-S Constrained & 0.905  &   0.605   &   0.868  &   0.713 & 0.734  \\
GPT-OSS F-S Implausible values  Constrained & 0.932  &   0.706   &   0.852  &   0.772 & 0.876 \\
MedLlama 0-S Constrained & 0.864 & 0.500 & 0.001 & 0.001 & 0.00 \\
MedLlama F-S Constrained &  0.780  & 0.198  &  0.202 & 0.200 & 0.620 \\
MedLlama F-S Implausible Constrained & 0.880 & 0.749 & 0.177 & 0.286 & 0.901 \\
MedGemma 0-S Constrained & 0.871  &   0.526   &   0.509 &   0.517 & 0.834   \\
MedGemma F-S Constrained &  0.856  &   0.476   &  0.599   &  0.531 & 0.838  \\
MedGemma F-S Implausible Constrained & 0.900 & 0.649 & 0.573 & 0.608 & 0.874 \\
Mistral 0-S  Constrained + RegEx & 0.896  &   0.628   &   0.568 &   0.596  & 0.908  \\
Mistral F-S  Constrained + RegEx &  0.910  &   0.768   &  0.485   &   0.594 & 0.906  \\
Mistral F-S Implausible values  Constrained + RegEx & 0.914  &   0.798   &   0.493  &   0.609 & 0.932 \\
Llama 0-S  Constrained + RegEx & 0.949 & 0.838 & 0.777 & 0.806 & 0.922 \\
Llama F-S  Constrained + RegEx & 0.930 & 0.753 & 0.714 & 0.733 & 0.916 \\
Llama F-S Implausible values  Constrained + RegEx & 0.880  &   0.754   &   0.176  &   0.286 & 0.903  \\
GPT-OSS 0-S  Constrained + RegEx & 0.948  &   0.858  &   0.739  &   0.794 & 0.930  \\
GPT-OSS F-S  Constrained + RegEx & 0.936  &   0.805   &   0.695  &   0.746 & 0.912 \\
GPT-OSS F-S Implausible values  Constrained + RegEx & 0.950 &   0.820   &   0.810  &   0.815 & 0.918 \\
MedLlama 0-S  Constrained + RegEx & 0.864 & 0.000 & 0.000 & 0.000 & N/A \\
MedLlama F-S  Constrained + RegEx &  0.842 & 0.333  &  0.165 & 0.221  & 0.751 \\
MedLlama F-S Implausible Constrained + RegEx & 0.880 & 0.754 & 0.176 & 0.286 & 0.903 \\
MedGemma 0-S  Constrained + RegEx & 0.877  &   0.552 &   0.499 &   0.524  & 0.847  \\
MedGemma F-S  Constrained + RegEx &  0.871  &   0.523 &  0.585 & 0.552 & 0.852  \\
MedGemma F-S Implausible  Constrained + RegEx & 0.903 & 0.669 & 0.566 & 0.614 & 0.881 \\
\midrule
Prompting Robustness - Second JSON Template &  & &  &  \\
\midrule
Mistral 0-S  & 0.915  &   0.648   &   0.799  &   0.714 &  0.893 \\
Mistral F-S  &  0.886  &   0.575   &  0.627   &   0.600 & 0.903 \\
Mistral F-S Implausible values &  0.806  &   0.300   &  0.323   &   0.311 & 0.046  \\
Llama 0-S &  0.948  & 0.769  &  0.872 & 0.817 & 0.885 \\
Llama F-S &  0.902  &  0.608  &  0.806  & 0.693 & 0.906 \\
Llama F-S Implausible values &  0.944  &  0.791  &  0.798  & 0.795 & 0.926  \\
GPT-OSS 0-S  & 0.968  &   0.877   &  0.886  &   0.881 & 0.941 \\
GPT-OSS F-S  & \textbf{0.968} &  0.884  &  0.876 &  0.880 & 0.940 \\
GPT-OSS F-S Implausible values & \textbf{0.968} &  \underline{0.885}   &  \textbf{0.881} &  \textbf{0.883} & \underline{0.943} \\
MedGemma 0-S  &0.933 & 0.910 & 0.559 & 0.692 & 0.927 \\
MedGemma F-S  & 0.916 & 0.794 & 0.512 & 0.623 & 0.922 \\
MedGemma F-S Implausible & 0.915 & 0.831 & 0.470 & 0.600 & 0.916 \\
\bottomrule
\end{tabular}
\label{tab:presence_absence_results}
\begin{tablenotes}
    \centering
    \item  0-S - zero-shot; Constrained - Constrained Generation with Guidance; F-S - few-shot; Implausible - adding implausible values to the few-shot prompt; RegEx- Regular Expressions. Settings with no extraction where not included in this table. The best results for each metric are shown in bold, while underline results represent the second best.
    \end{tablenotes}
\end{table*}

\end{document}